\documentclass[letterpaper, 10 pt, conference]{ieeeconf}
\IEEEoverridecommandlockouts                              % This command is only needed if 
\usepackage{amsmath,amssymb,amsfonts}
\usepackage{graphicx}
\usepackage{textcomp}
\usepackage[table]{xcolor}
\usepackage{booktabs}
\usepackage{makecell}
\usepackage{multirow}
\usepackage{caption}
\usepackage{subcaption}
\usepackage{algpseudocode}
\usepackage{algorithm}
\usepackage{float}
\usepackage{diagbox}
\usepackage{lineno}
\usepackage{setspace}
\usepackage{ulem}
\usepackage{xcolor}
\usepackage{moreverb,url}
\usepackage{amssymb}

\usepackage[colorlinks,bookmarksopen,bookmarksnumbered,citecolor=red,urlcolor=red]{hyperref}
\hypersetup{
    colorlinks,
    linkcolor={red!30!black},
    citecolor={blue!30!black},
    urlcolor={blue!30!black}
}
\title{\LARGE \bf
Vision-based robot manipulation of transparent liquid containers in a laboratory setting
}
\author{Daniel Schober$^{1}$, Ronja Güldenring$^{1}$, James Love$^{2}$ and Lazaros Nalpantidis$^{1}$% <-this % stops a space
\thanks{The authors would like to thank Novo Nordisk for providing all the hardware equipment for the developed robotic setup and the team of Automation and Process Optimization for their invaluable support.}% <-this % stops a space
\thanks{$^{1}$Technical University of Denmark, Kongens Lyngby, Denmark
        {\tt\small \{s212599, ronjag, lanalpa\}@dtu.dk}}%
\thanks{$^{2}$Automation and Process Optimization, Novo Nordisk, Målov, Denmark
        {\tt\small }}%
}

\begin{document}

\maketitle
\thispagestyle{empty}
\pagestyle{empty}

%%%%%%%%%%%%%%%%%%%%%%%%%%%%%%%%%%%%%%%%%%%%%%%%%%%%%%%%%%%%%%%%%%%%%%%%%%%%%%%%
\begin{abstract}
Laboratory processes involving small volumes of solutions and active ingredients are often performed manually due to challenges in automation, such as high initial costs, semi-structured environments and protocol variability. In this work, we develop a flexible and cost-effective approach to address this gap by introducing a vision-based system for liquid volume estimation and a simulation-driven pouring method particularly designed for containers with small openings. We evaluate both components individually, followed by an applied real-world integration of cell culture automation using a UR5 robotic arm. Our work is fully reproducible: we share our code at at \url{https://github.com/DaniSchober/LabLiquidVision} and the newly introduced dataset \textit{LabLiquidVolume} is available  at \url{https://data.dtu.dk/articles/dataset/LabLiquidVision/25103102}.
\end{abstract}

\section{Introduction}
% General introduction to the field of laboratory assistants
In life science and chemistry labs, robots are already used in highly standardized processes that deal with large amounts of samples and require high repeatability. However, experiments with small amounts of solutions and active ingredients samples are still mostly executed manually by specialists in the lab. Among the reasons for this automation gap are the high initial investment costs, semi-structured nature of R\&D labs, high variability in research protocols, and the current lack of lower-cost interim labor-saving automation options, as attested in \cite{Wolf2019AAutomation, Groth2017IndicatorsAnalysis}. 

In this work, we introduce a flexible, modular and cost-effective robotic system to allow the flexible automation of small-scale laboratory experiments involving manipulation of liquid containers. Our robot is guided by computer vision to estimate liquid volumes and it adopts a simulation-driven pouring approach particularly designed for containers with small openings.

\begin{figure}[!tb]
\centering
\begin{subfigure}{.99\linewidth}
\centering
    \includegraphics[width=\linewidth]{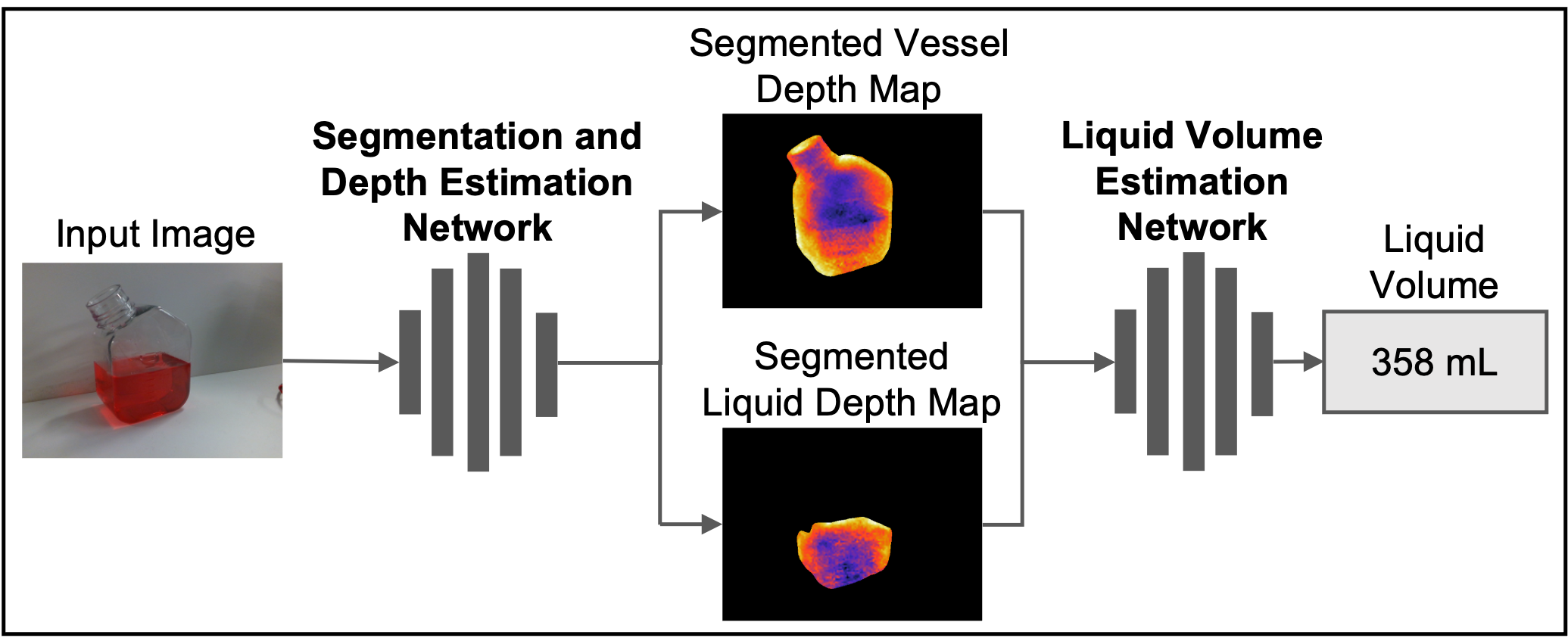}
\caption{}
\label{fig:TwoStep}
\end{subfigure}\\
\begin{subfigure}{.99\linewidth}
\centering
    \includegraphics[width=\linewidth]{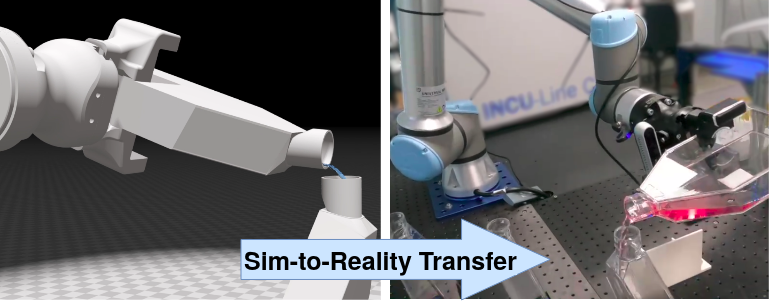}
\caption{}
\label{fig:kp_target}
\end{subfigure}
\caption{We present a robotic setup for cell culture automation using exclusively existing laboratory equipment. (a) First, we propose a two-step model architecture to estimate the liquid volume in transparent laboratory containers using a single RGB image. (b) Secondly, from a large pool of simulated pouring trajectories, the simulation that approximates the real-world state best is performed on the UR5. Hereby, the pouring strategy is constrained to rotate around the liquid exit point which is crucial for receiving containers with small openings such as cell culture flasks.}
\label{fig:intro_fig}
\end{figure}

Our proposed vision-based liquid volume estimation method, illustrated in Fig.~\ref{fig:TwoStep}, uses a novel two-step Convolutional Neural Network (CNN) architecture. In the first step, a single RGB input image is processed by the first CNN to predict the segmentation and depth of the transparent container and the contained liquid. For the training of the first step, we benefit from existing datasets targeting transparent containers such as \textit{TransProteus} \cite{Eppel2022PredictingDataset} and \textit{Vector-LabPics} \cite{Eppel2020ComputerDataset}. These intermediate predictions are further processed by a second CNN to give an estimate of the liquid volume in the container. For the training of the second network, we create a new dataset \textit{LabLiquidVolume} with 5,451 RGB images of transparent laboratory containers manually labeled with the liquid volume inside the container. Our two-step architecture allows us to keep the size of the newly introduced dataset for training the second step relatively small, as we are able to build upon the knowledge distilled from the publicly available datasets during the first step.

Additionally, we have developed a strategy for the robot to pour from containers with small openings, by confining the robot arm to rotate around a fixed liquid exit point. From a large pool of simulated pouring trajectories applicable for small openings, the simulation that approximates the real-world state best is selected to be executed in the real-world automation setup; this is illustrated in Fig.~\ref{fig:kp_target}.
Pouring a specific amount of liquid into a container with a small opening is an important and frequently reoccurring manipulation task for robots in laboratory environments. Hereby we set a focus on the flexible usage, i.e. the pouring procedure can be performed in different experimental setups with visual and auditory variations.

% Mention fully integrated platform.
We showcase our developed system on automated cell culture processes, which include various pouring processes such as media changing and passaging cells. We integrate a UR5 robotic arm to fully automate the cell culture process using exclusively existing laboratory equipment. Videos of the automated cell culture process including the different autonomous pouring steps can be seen here: \url{https://github.com/DaniSchober/LabLiquidVision}. 

Our contributions in this work can be summarized as follows.
\begin{itemize}
    \item We propose a novel two-step CNN architecture to perform liquid volume estimation for transparent laboratory containers. The two-step nature of our architecture has the advantage that we can benefit from a large pool of relevant existing datasets in the first stage and predict intermediate properties such as the depth and segmentation of liquid and vessel.
    \item We publish a newly created dataset \textit{LabLiquidVolume} with 5,451 images to enable the training of actual liquid estimation in the second step of our proposed two-step CNN architecture. To the best of our knowledge, it is the first real-world dataset including labels for the liquid volume in the transparent container. The data can be downloaded at \url{https://data.dtu.dk/articles/dataset/LabLiquidVision/25103102}.
    \item We develop a new pouring strategy for containers with small openings by confining the robot arm to rotate around a fixed liquid exit point, inspired by human execution, and showcase the operation of the entire system on a physical robotic setup.
\end{itemize}

\section{Related Work}
Laboratory processes at small scale are often performed manually because the automation of those processes causes high initial costs and includes protocol variability which in most cases does not pay off. Robots constitute a potential solution for the automation of small-scale and variable laboratory processes because they can be deployed for one-to-one replacement of manual labor without the need for any additional specialized equipment as is the case for the automation of large-scale processes. Hereby, different kinds of robots can be relevant such as mobile robots~\cite{Burger2020AChemist, Kleine-Wechelmann2023DesigningLaboratory} (usually including a moveable arm), stationary robots~\cite{Yoshikawa2022ChemistryPlanning, LabmanAutomationLtd2023MultiDose:System}, and fluidic robots~\cite{Steiner2019OrganicLanguage}. In our work, we aim to modularize the pouring process using standard laboratory containers with small openings, which is relevant for various cell culture procedures and life science laboratory processes. It includes the automated volume estimation of liquids in transparent containers as well as a pouring strategy complying with the small openings of the containers.

\subsection{Vision-based Liquid Volume Estimation}
The following publications have been identified to perform vision-based liquid volume estimation in transparent containers. Mottaghi et al.~\cite{Mottaghi2017SeeContent} formulate the task as a classification problem and define discrete classes for ranges of the container sizes (50, 100, 200, 300, 500, 750, 1000, 2000, 3000 mL, and above) as well as liquid fillings in percentage (0\%, 33\%, 50\%, 66\%, 100\% of the container volume). The classification CNN is trained on single RGB images. Zhu et al.~\cite{Zhu2022Visual-TactileGrasping} enrich the input RGB image with the measurement of a customized tactile sensor and train a multi-modal CNN which predicts the liquid volume with an error of 2 mL for volumes of 40-80 mL. Cobo et al.~\cite{Cobo2022ArtificialImages} determine the wine volume on a single-view RGB image with a mean average error lower than 10 mL for volumes of 50-300 mL.

The above-presented approaches have been applied to domains outside laboratory settings. In laboratory settings, only the detection of materials (including liquids) in transparent containers has been heavily researched by Eppel et al.. Initially, they investigated traditional image processing methods such as edge detection~\cite{Eppel2014ComputerApplications} and graph-cuts~\cite{Eppel2016TracingApproach}. Recently, they have focused on liquid detection using CNNs trained on a relatively small real-world dataset \textit{Vector-LabPics}~\cite{Eppel2020ComputerDataset} as well as a synthetically generated larger dataset \textit{TransProteus}~\cite{Eppel2022PredictingDataset}. 
\\\\
In our work and to the best of our knowledge, we present the first vision-based liquid volume estimation approach developed for laboratory settings. We take advantage of the publicly available laboratory datasets \textit{Vector-LabPics}~\cite{Eppel2020ComputerDataset} and \textit{TransProteus}~\cite{Eppel2022PredictingDataset}. We apply a two-step deep learning architecture with two consecutive models. The first model generates a segmentation and depth map, followed by a regression model to estimate the actual liquid volume.

\subsection{Autonomous Pouring}
Existing pouring strategies tackle pouring into wide openings, like e.g. in kitchen assistant tasks, leading to moving liquid exit points while pouring. The majority of pouring strategies~\cite{Kennedy2019AutonomousContainers, Do2018AccurateCamera, Schenck2017VisualLiquids, Huang2021RobotGeneralization, Guevara2017AdaptableSimulation, Lopez-Guevara2020StirActions} perform a rotation around the wrist of the robot, while others use more advanced approaches for the trajectory planning \cite{Babaians2022PourNet:Learning, Pan2016RobotLiquids}.
\\\\
For our laboratory automation, we confine the pouring strategy to rotate around a fixed liquid exit point, which is especially relevant for pouring in small openings, such as cell culture flasks. Finally, the constrained pouring strategy is learned using approximate simulation as in \cite{Guevara2017AdaptableSimulation, Kennedy2019AutonomousContainers}. The final robot movement is performed according to the simulated movement that approximates the real-world scenario closest.

\section{Methods}
\subsection{Vision Model for Liquid Volume Estimation}
The objective is to predict the volume inside transparent laboratory vessels based on a single RGB image. Inspired by Graikos et al.~\cite{Graikos2020SingleNetworks} and illustrated in Fig.~\ref{fig:TwoStep}, we propose a two-step deep learning architecture consisting of two sub-models: (1) the \textit{Segmentation and Depth Estimation (SDE) Network} and (2) the \textit{Liquid Volume Estimation (LVE) Network}.
\\\\
The SDE Network is based on the work presented by Eppel et al.~\cite{Eppel2022PredictingDataset}. In contrast to Eppel et al.~\cite{Eppel2022PredictingDataset}, we take advantage of a constant camera setup with known camera parameters, enabling the prediction of a 2D depth map instead of a 3D XYZ map. For the architecture, a modified DeepLab-v3 is used, which takes as input an RGB image and outputs both a segmentation map and a depth map. Note that the segmentation map covers three classes: vessel, vessel opening, and liquid.
\\\\
The LVE Network processes the output of the SDE Network to predict the vessel's liquid volume in mL. The architecture is a simple down-sampling Convolutional Neural Network, consisting of five convolutional layers, followed by four fully connected layers.  

\subsection{Pouring in Small Openings}
\label{sec:PouringSmallOpenings}
\begin{figure}
\centering
    \includegraphics[width=1.0\linewidth]{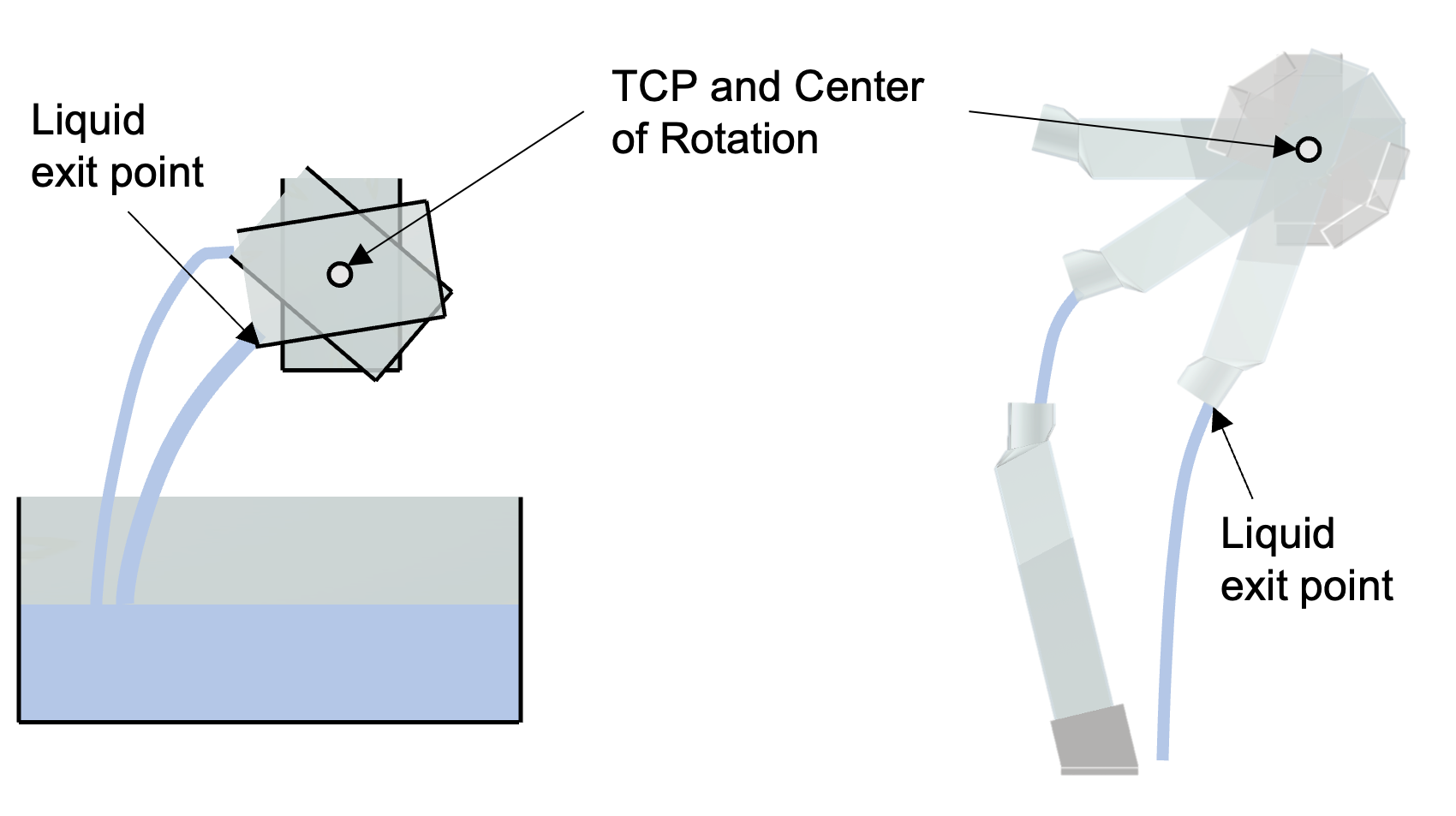}
\caption{Visualization of the issue with pouring movements only considering a wrist rotation. Left: Successful pouring only using rotation because of a receiving container with a large opening surface and a pouring container with a small height. Right: Liquid is spilled when only rotating a cell culture flask around the TCP.}
\label{fig:Pouring_Bad_Movement}
\end{figure}

Robotic pouring with rotation of the wrist is not applicable for containers with small openings, such as cell culture flasks, because the exit point of the liquid is constantly changing. As illustrated in Fig. \ref{fig:Pouring_Bad_Movement}, this would result in a large amount of spillage. In research laboratories, particularly where liquids can pose a safety hazard, minimizing the volume of spills is of critical importance.

We propose a method for pouring liquids into containers with a narrow opening inspired by human movements. It mimics the human approach of initially aligning the pouring container's opening with the receiving container's opening, followed by rotating the pouring container around the fixed liquid exit point. In this method, the tool center point $TCP$ undergoes both positional and orientational changes during the movement. The rotation angle ($\theta$) around the liquid exit point, referred to as the center of rotation ($CoR$), determines the coordinates of the $TCP$ in the XY plane. Depending on the pouring container, the pouring path is calculated based on the distance between the center of rotation and tool exit point $l = |TCP_{start} - CoR|$, the angle  $\beta$ between them, and the rotation at the start $\alpha$. $TCP_x$ and $TCP_y$ for a specific $\theta$ can then be calculated as follows:

For $\theta \leq \alpha_{start}$:
\begin{align*}
    TCP_x(\theta) &= TCP_{x,start} - l \cos(\alpha_{start}) + l \cos(\alpha_{start} - \theta)\\ 
    TCP_y(\theta) &= TCP_{y,start} + l \sin(\alpha_{start}) - l \sin(\alpha_{start} - \theta)
    \label{eq:TCP4}
\end{align*}

\begin{figure}
\centering
    \includegraphics[width=1.0\linewidth]{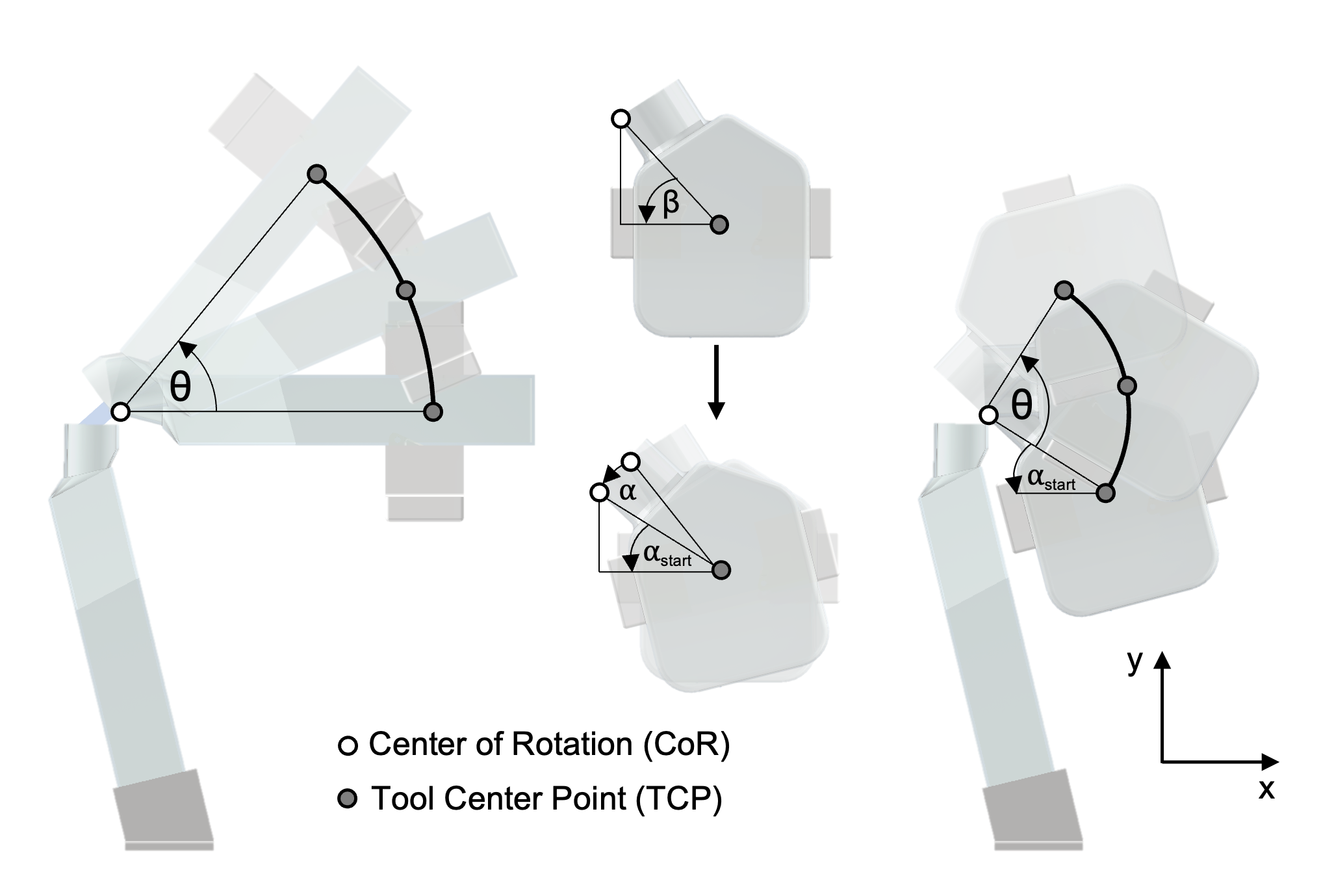}
\caption{Visualization of the pouring movement for two exemplary laboratory containers. $\theta$ represents the pouring angle, $\beta$ the angle between the $TCP$ and the $CoR$ of the bottle in the horizontal position, $\alpha$ the rotation of the bottle to the same angle as the receiving flask ($\alpha=14.5°$), and $\alpha_{start}$ the start angle between the $TCP$ and the $CoR$. Left: Pouring with the cell culture flask. Center: Explanation of the start position of the bottle. Right: Pouring with the media/washing solution bottle.}
\label{fig:Cor_TCP}
\end{figure}

For $\theta > \alpha_{start}$:
\begin{align*}
    TCP_x(\theta) &= TCP_{x,start} - l \cos(\alpha_{start}) + l \cos(\theta - \alpha_{start})\\ 
    TCP_y(\theta) &= TCP_{y,start} + l \sin(\alpha_{start}) + l \sin(\theta - \alpha_{start})
\end{align*}
The proposed pouring movement is illustrated in Fig. \ref{fig:Cor_TCP}.

\subsection{Simulation-to-Reality Transfer of Liquid Pouring}
Inspired by Kennedy et al.~\cite{Kennedy2019AutonomousContainers} and Lopez-Guevara et al.~\cite{Guevara2017AdaptableSimulation}, a simulation is used to pre-evaluate various pouring movements and identify the optimal robot movement for specific input parameters in real-world scenarios. The main advantage of using pre-simulated pouring results is the reduction of numerous time-consuming real pouring attempts, allowing for more efficient experimentation and adaptation. Moreover, a detailed target volume precision achieved in simulations is possible because of the possibility of varying various pouring parameters.

We use $\mathbf{p} \in \mathbf{P}$ to denote a set of parameters, $\mathbf{p}^{(i)}$ to denote the $i^{th}$ dimension of the set of parameters, and $\mathbf{p}^{*}$ to represent an optimal set of parameters.
Hence, the combination of the remaining 2D \textit{action parameter} space and the 2D \textit{simulation parameter} space results in a reduced 4D parameter space $\mathbf{p}$. This consists of the pouring container $\mathbf{p}^{(0)} = c$, the start volume in the pouring container $\mathbf{p}^{(1)} = V_{start}$, the stop angle $\mathbf{p}^{(2)} = \theta _{stop}$, and the stop time $\mathbf{p}^{(3)} = t_{stop}$. The angular velocity and start position were kept constant.

The start volume $C_{start}$, the received volume $C_{received}$, and the spilled volume $C_{spill}$ of each scene were stored. Based on the given real-world parameters, which are the pouring container, the start volume, and the desired volume to be poured $V_{goal}$, the best-fitting simulated scene with an optimal set of parameters $\mathbf{p}^{*}$ can then be chosen by iterating through the results of the simulation and selecting the one with the lowest cost:

\begin{equation}
C(\mathbf{p}) = C_{start} + C_{received} + C_{spill}
\label{eq:Cost}
\end{equation}

$C_{start}$ is the magnitude of the difference between the start volume in the simulation scene and the start volume in the real-world scenario. $C_{received}$ is the magnitude of the difference between the received volume in the receiving container in the simulation and the desired real-world volume to be received. $C_{spill}$ is the volume of spilled liquid in the simulation.

\section{Experimental Setup}
\subsection{Cell Culture Robotics Setup}
\subsubsection{Real-World}
\begin{figure}[!htb]
\centering
    \includegraphics[width=1.0\linewidth]{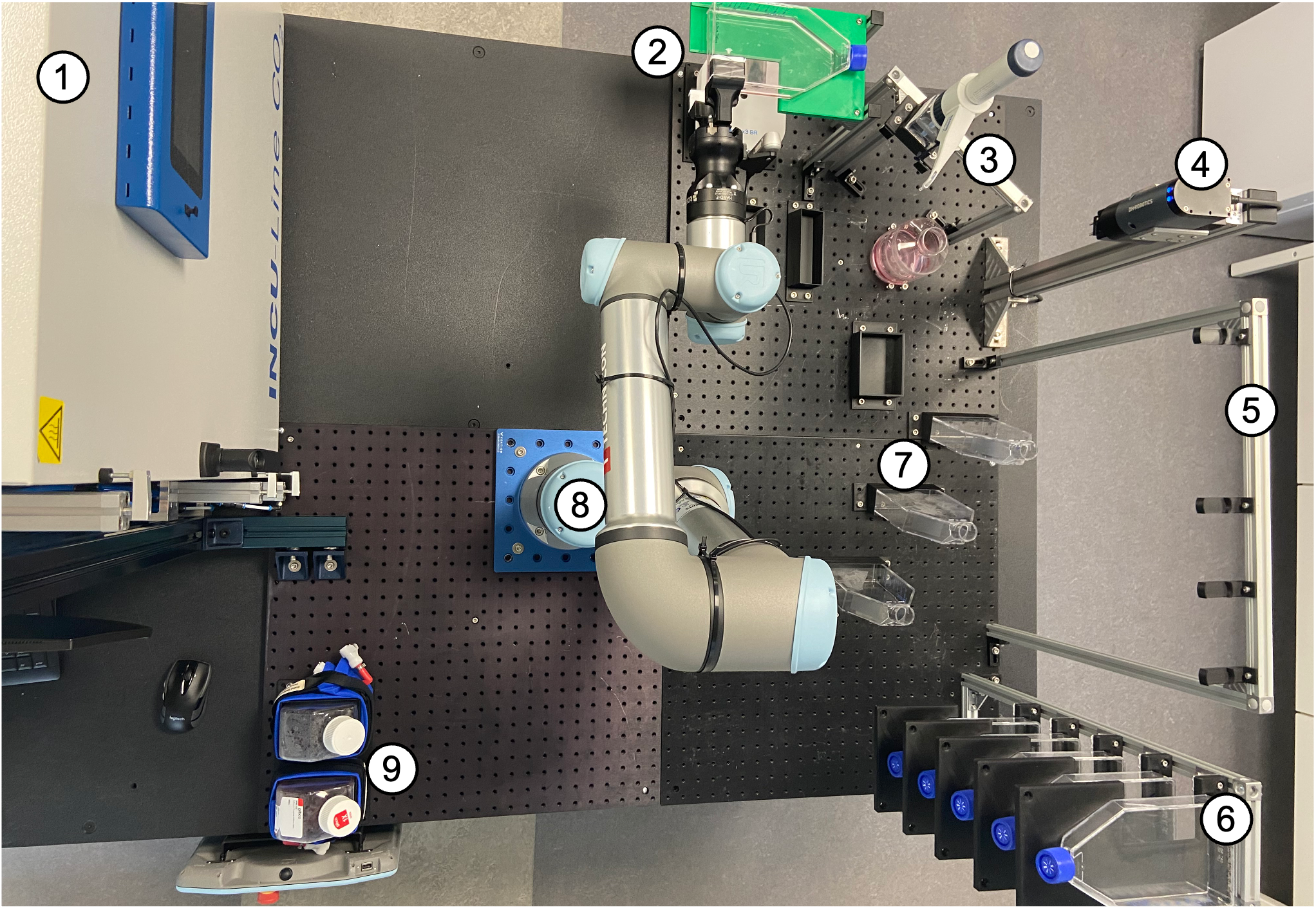}
\caption{Overview of the system prototype for cell culture automation. (1) Automated incubator. (2) Microscope. (3) Trypsin unit. (4) Capper/Decapper. (5) Lid holders. (6) Flask storage. (7) Flask holder for pouring. (8) UR5e with 3D-printed gripper fingers. (9) Heating and cooling of liquids unit.}
\label{fig:Prototype}
\end{figure}

In the planning and design phase of the system, key considerations were modularity, simplicity, and the integration of existing laboratory equipment. It includes a UR5e, a CytoSMART Lux3 BR microscope, an Intel RealSense D415, a  VWR INCU-Line ILCO 180 Premium CO\textsubscript{2} incubator, and a thermoelectric chiller. Only consumables utilized in the manual cell culture process carried out by laboratory labor are used to enable seamless system integration. The setup is visualized in Fig. \ref{fig:Prototype}. All the devices are either connected to the UR control box or directly to the PC. Similar to the manual process, the autonomous process is split into three high-level workflows. 
\begin{itemize}
    \item Workflow A: Analyzing cell growth.
    \item Workflow B: Changing media.
    \item Workflow C: Passaging.
\end{itemize}

The system should be able to execute the three workflows independently, depending on the user input or the scheduled plan. To enhance reusability and simplicity, the steps to achieve completion of the workflows are structured into workflow modules. Pouring is required for the removal of spent media, adding and discarding of the washing solution, adding of fresh media, and splitting of the cell culture into three empty flasks. Since only a very low amount of dissociation reagent such as trypsin (approximately 0.5 mL per 10 cm\textsuperscript{2}) needs to be added to the cell culture flask, pouring is unsuitable. Instead, a bottle dispenser (1-10 mL) was attached to a trypsin bottle to allow vertical insertion of the reagent into the flask. All real-world pouring experiments were executed on that setup.
\subsubsection{Simulation}
\begin{figure}
\centering
\includegraphics[width=1.0\linewidth]{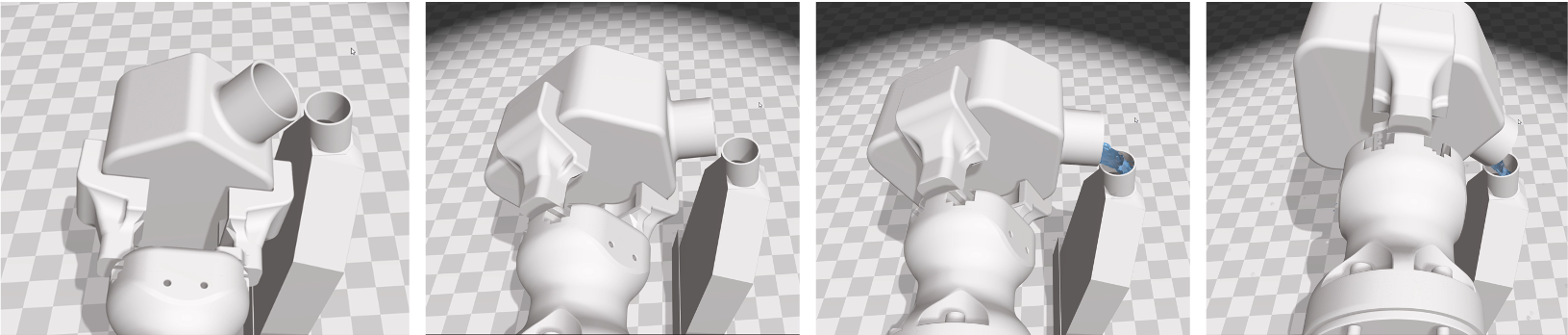}
\caption{An exemplary simulation scene of the media bottle executing the new pouring movement.}
\label{fig:Pouring_seq}
\end{figure}

Just like Kennedy et al.~\cite{Kennedy2019AutonomousContainers} and Lopez-Guevara et al.~\cite{Guevara2017AdaptableSimulation}, we used the particle-based simulation library NVIDIA Flex for the simulation of pouring movements. Two scenes are created to simulate the two main liquid pouring movements required for cell culture automation. One for pouring from a cell culture flask, and one for pouring from a media bottle, both into another standing flask. The pouring movement is executed as explained in \ref{sec:PouringSmallOpenings}. An exemplary simulation scene is shown in Fig. \ref{fig:Pouring_seq}. In total, 6,805 pouring movements were simulated.

\subsection{Volume Estimation of Liquids}
\subsubsection{Vision Datasets}
In this work, we consider two recently published datasets containing transparent objects with liquid content, namely \textit{TransProteus}~\cite{Eppel2022PredictingDataset} and \textit{Vector-LabPics}~\cite{Eppel2020ComputerDataset}. Furthermore, a new dataset that will be called \textit{LabLiquidVolume} was created for this work.

The \textit{TransProteus}~\cite{Eppel2022PredictingDataset} dataset consists of more than 50,000 synthetic images with annotated 3D models of transparent vessels and their content, including segmentation masks and material properties such as color, transparency, reflectance, and roughness. For this work, only transparent vessels with liquid material are considered, which constitutes \(\approx\) 14,500 samples.

The \textit{Vector-LabPics}~\cite{Eppel2020ComputerDataset} dataset comprises 7,900 real-world images with pixel-wise annotation of vessels and the containing material. The images capture materials in diverse stages and procedures observed in transparent vessels within chemistry labs, medical labs, hospitals, and other relevant settings.

\begin{figure}[!htb]
\centering
    \includegraphics[width=1.0\linewidth]{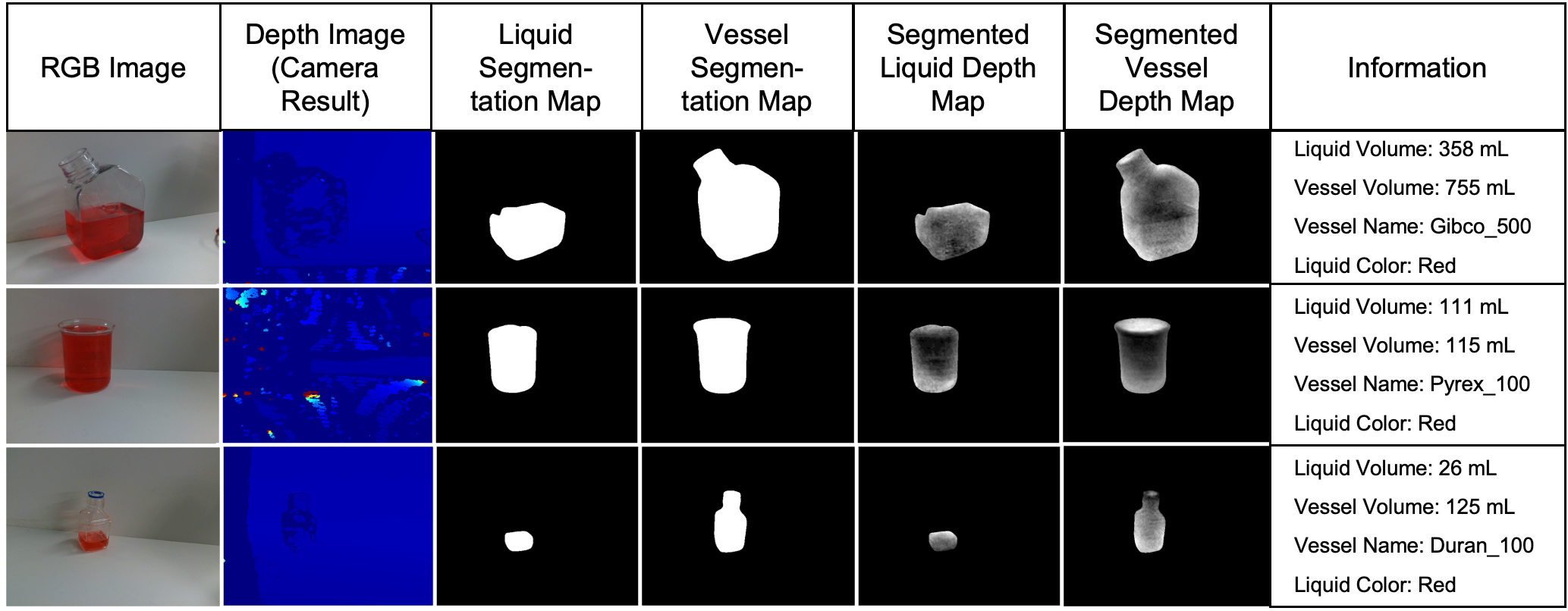}
\caption{Visualization of the content of three samples of the generated \textit{LabLiquidVolume} dataset.}
\label{fig:LabLiquidVolume}
\end{figure}

The \textit{LabLiquidVolume} dataset includes 5,451 images of liquids in laboratory containers. The images were taken using an Intel RealSense D415 camera in different environments in the automation laboratory and various research laboratories at the Novo Nordisk R\&eD site in M\r{a}l\o{}v, Denmark. The ground truth of the liquid volume was measured using a Mettler Toledo XSR2002S balance with an accuracy of ± 0.5 mL. Twelve of the most common research laboratory containers, including the consumables used in cell culture processes, were selected. Various liquid volumes (3 - 600 mL),
distances to the containers (50 - 600 mm), backgrounds, and camera angles were used. However, all images were taken from above so that the surface of the liquid is visible.
The trained SDE model was used to generate the segmentation masks and depth maps automatically.
Additionally, for each container, 10 images were manually annotated with a pixel-wise mask for vessel and liquid.  Fig.~\ref{fig:LabLiquidVolume} presents example data points of the dataset.

\subsubsection{Model Training Settings}
\label{sec:model_training_settings}
The SDE network is trained for 75 epochs on the relevant subset of the \textit{TransProteus} dataset and the \textit{Vector-LabPics} dataset. Hereby, images from the \textit{Vector-LabPics} are sampled with \(33\%\), and the depth loss is set to zero, because no depth ground truth is available for those images. Image augmentation for training data includes resizing, centered cropping, Gaussian blurring, decoloring, and darkening. An Adam optimizer was used with an initial learning rate of $1\times 10^{-5}$ and a weight decay of $4\times 10^{-5}$. The learning rate is adjusted every five epochs based on the average loss. If the average loss does not decrease for ten consecutive epochs, the learning rate is reduced by 10\%. The batch size was set to 6.  The loss for each segmentation mask is calculated using the standard per-pixel cross-entropy function. 
The loss for each depth map is calculated only for the region of the object as given by the specific ground truth mask. The scale-invariant depth loss, as introduced by Eigen et al. \cite{Eigen2014DepthNetwork}, is applied. The final model is evaluated on the subset of the official \textit{TransProteus} test set, which contains 1,500 samples for objects containing liquid.

\begin{figure}[!htb]
\centering
    \includegraphics[width=1.0\linewidth]{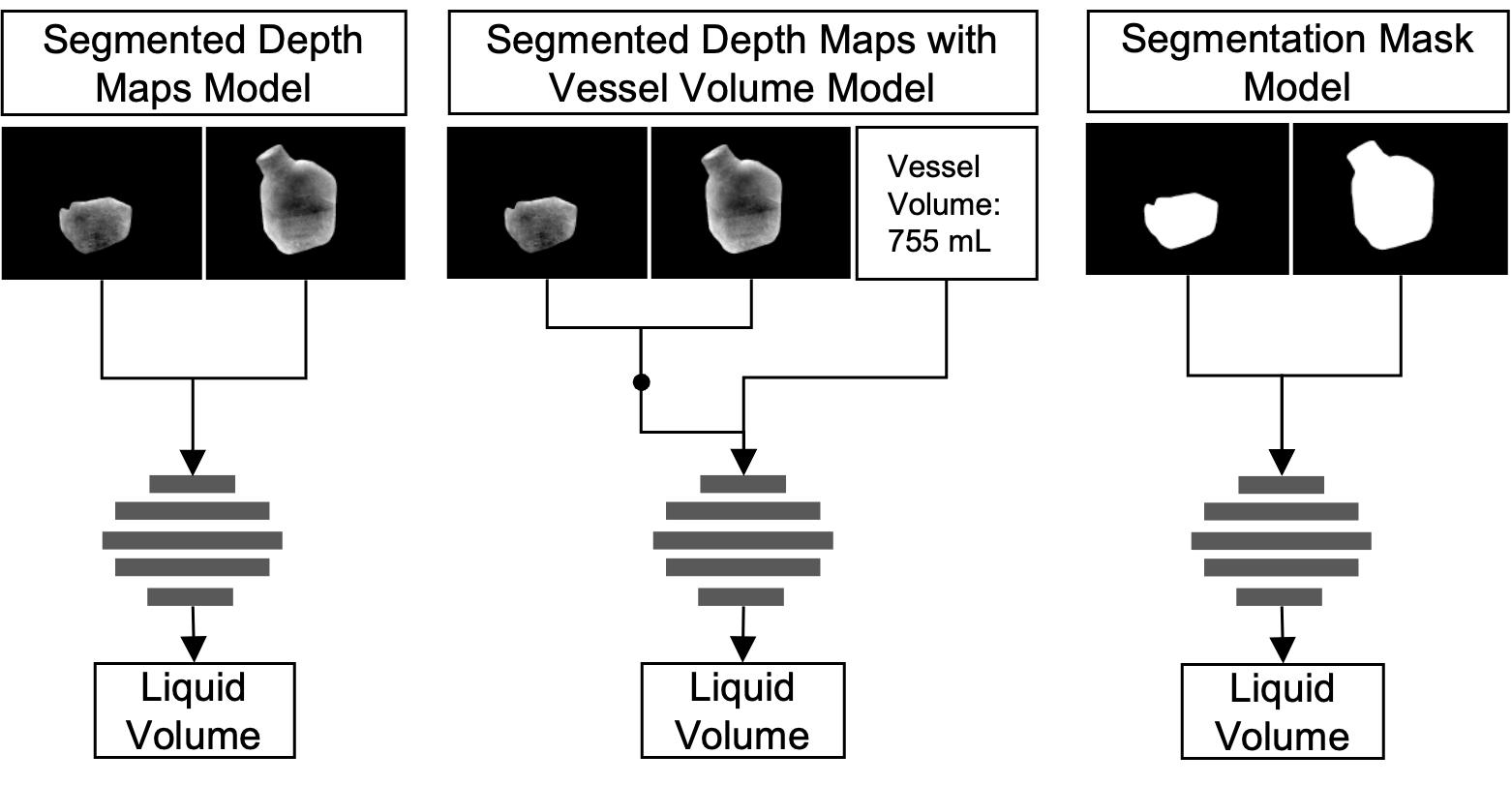}
\caption{Overview of the three input variants for liquid volume estimation network, trained on the \textit{LabLiquidVolume} dataset.}
\label{fig:ModelComparison}
\end{figure}

The LVE network is trained on 80\% of the \textit{LabLiquidVolume} dataset. Hereby, the impact of different input information is investigated, as illustrated in Fig.~\ref{fig:ModelComparison}.
\begin{itemize}
    \item[(1)] Vessel depth map \& liquid depth map
    \item[(2)] Vessel depth map \& liquid depth map \& vessel volume
    \item[(3)] Vessel segmentation map \& liquid segmentation map
\end{itemize}

For the second variant, the vessel volume is processed by a fully connected layer before being concatenated with the segmented depth maps. All variants are trained for 200 epochs with an Adam optimizer, a learning rate of 0.001, a batch size of 8, and a dropout rate of 0.2. For the loss function, we apply the Mean Squared Error (MSE).

\subsubsection{Evaluation Metrics}
For the evaluation of the SDE model only, we use the widely used Intersection over Union ($IoU$) to measure the segmentation performance and the $scale-invariant \; log \; RMSE$ suggested by Eigen et al. \cite{Eigen2014DepthNetwork} to measure the performance of the depth map prediction.
For the evaluation of the full model, three different regression metrics are used: The root-mean-squared error $RMSE$, the mean absolute percentage error $MAPE$, and the $R^2$ score (coefficient of determination).

\section{Experimental Results}
\subsection{Vision Model Performance}
For the SDE model, results are on par with the ones from Eppel et al.~\cite{Eppel2022PredictingDataset}. On the test subset, we achieve a mean $IoU$ of 0.92 (Vessel: 0.96, Vessel Opening: 0.94, Liquid: 0.85) for the segmentation performance and a \textit{scale -- invariant log RMSE} of 0.031 for the depth estimation. The segmentation performance was additionally evaluated on the 120 manually annotated images of the \textit{LabLiquidVolume} dataset. The performance on the real-world  \textit{LabLiquidVolume} images drops with a mean $IoU$ of 0.90 (Liquid: 0.87, Vessel: 0.92). Furthermore, we can report a gap between glass and plastic objects for liquid segmentation. While the liquid region inside glass vessels is predicted with an $IoU$ of 91.3\%, it is significantly lower for liquids inside plastic objects with an $IoU$ of 84.3\%.

\begin{figure}[!htb]
\centering
    \includegraphics[width=1.0\linewidth]{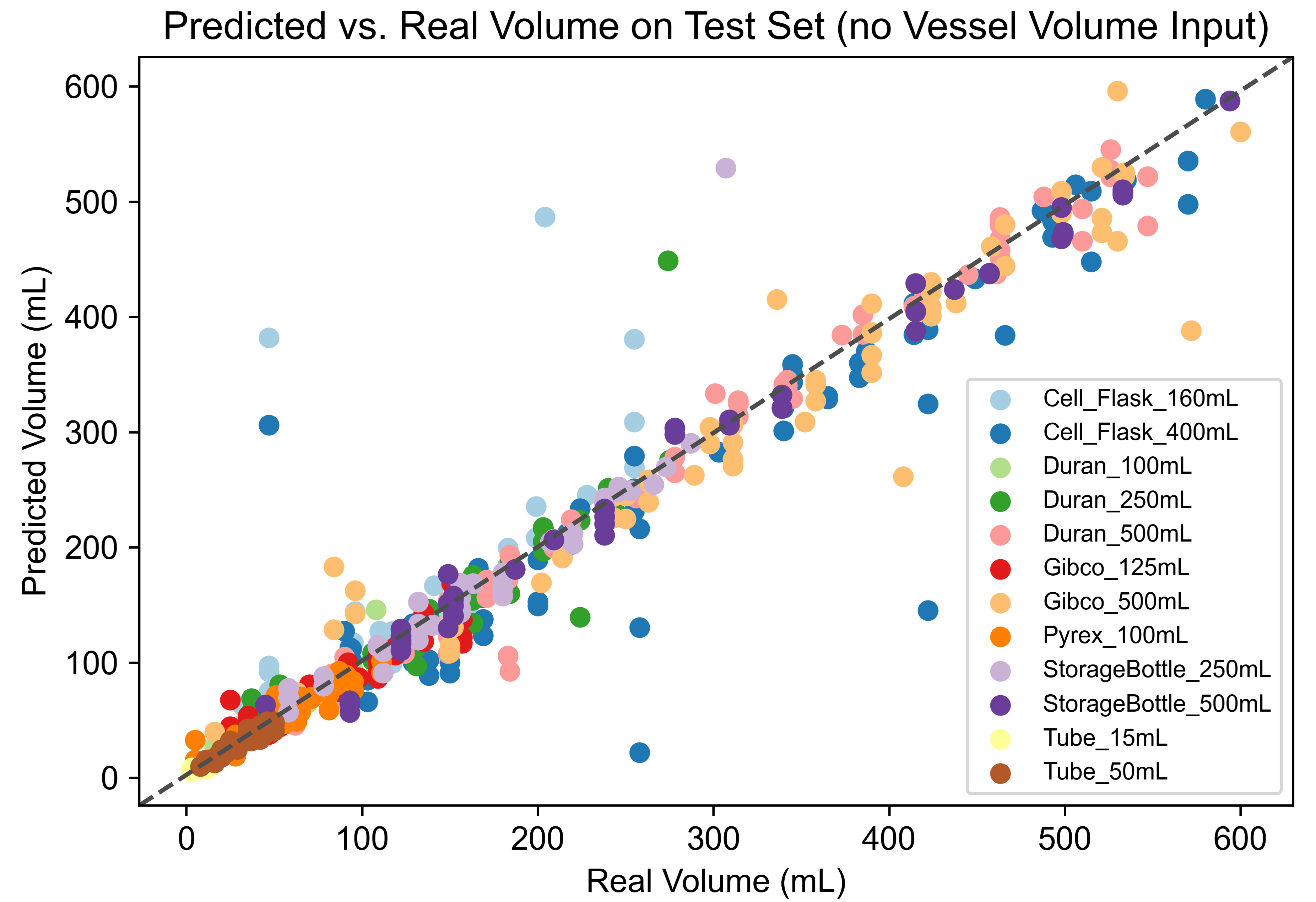}
\caption{Real and predicted liquid volumes for images from the \textit{LabLiquidVolume} test set for the model trained only on the segmented depth maps.}
\label{fig:Scatter_Volume_no_vessel}
\end{figure}

For the full model, we use the trained SDE model presented above and combine it with the three variants for the LVE model as described in section~\ref{sec:model_training_settings}. The results for the three variants are presented in Table~\ref{tab:Results_4_models} and indicate that the input combination of segmented depth maps and vessel volume model outperforms their competitors. Omitting depth information entirely and solely relying on liquid and vessel segmentation masks as input significantly degrades the model's performance.
\begin{table}[!htb]
\small
\centering
\setlength{\tabcolsep}{5pt} % Adjusted column separation
\caption{Performance of the three variants for the LVE model combined with the same SDE model.}
\begin{tabular}{@{}lcccc@{}}
\toprule
Model Input                            & \begin{tabular}[c]{@{}c@{}}RMSE\\ Training\\ (mL)\end{tabular} & \begin{tabular}[c]{@{}c@{}}RMSE\\ Testing\\ (mL)\end{tabular} & \begin{tabular}[c]{@{}c@{}}MAPE\\ Testing\\ (\%)\end{tabular} & \begin{tabular}[c]{@{}c@{}}$R^2$\\ Testing\end{tabular} \\ \midrule
Segmented Depth Maps                   & 13.00                                                              & 37.90                                                             & 16.96                                                             & 0.94                                                      \\
\makecell{Segmented Depth Maps\\ \& Vessel Volume} & 4.63                                                               & 17.83                                                             & 9.39                                                              & 0.99                                                      \\
Segmentation Masks                     & 32.94                                                              & 53.92                                                             & 24.21                                                             & 0.88                                                      \\ \bottomrule
\end{tabular}
\label{tab:Results_4_models}
\end{table}
However, in the further course of the experimental section, we will consider the LVE model that takes segmented depth maps only as input, because it allows the manipulation of vessels of unknown vessel volume and is, therefore, more flexible. Its distribution of real and predicted liquid volumes is shown in Fig. \ref{fig:Scatter_Volume_no_vessel}.

We found a moderate negative correlation coefficient of -0.69 between the $IoU$ for liquid segmentation and the $MAPE$ of liquid volume estimation, indicating that better liquid segmentation performance was linked to more accurate volume predictions.

\subsection{Pouring Performance}
\subsubsection{Simulation}
\begin{figure}[!htb]
\centering
\includegraphics[width=1.0\linewidth]{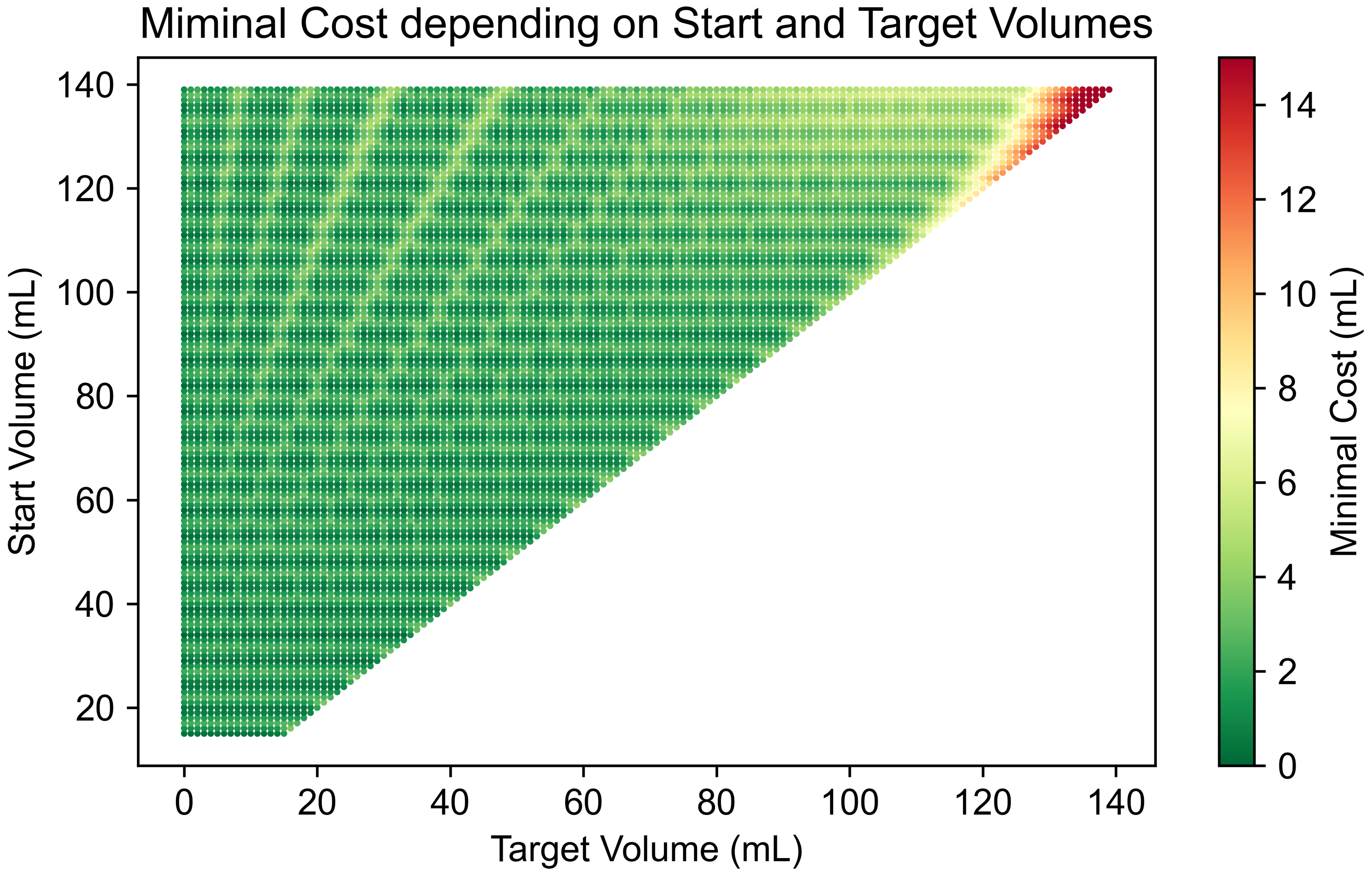}
\caption{Minimal costs for pours with a cell culture flask given a defined start and target volume with step sizes of 1 mL.}
\label{fig:Cost_map_flask}
\end{figure}

We perform a cost analysis by creating sets of every possible target volume for every possible start volume between 0 mL and the maximum of the received volumes with a step size of 1 mL each. For the pouring with the cell culture flask, this results in 9,750 possible combinations of start and target volumes. Fig.~\ref{fig:Cost_map_flask} shows the minimum cost depending on start and target volumes. The lowest cost among the combinations is 0.05 mL, and the highest is 16.5 mL. The mean cost for the set is 2.3 mL. For start volumes below 100 mL, the cost is smaller than 7.5 mL for every possible target volume, with a mean cost of 1.8 mL. The cost increases significantly for high start volumes with target volumes above 90 mL. For target volumes below 105 mL, a simulated pour can be found without any spilled liquid.

\begin{figure}[!htbp]
  \centering
  \includegraphics[width=1\linewidth]{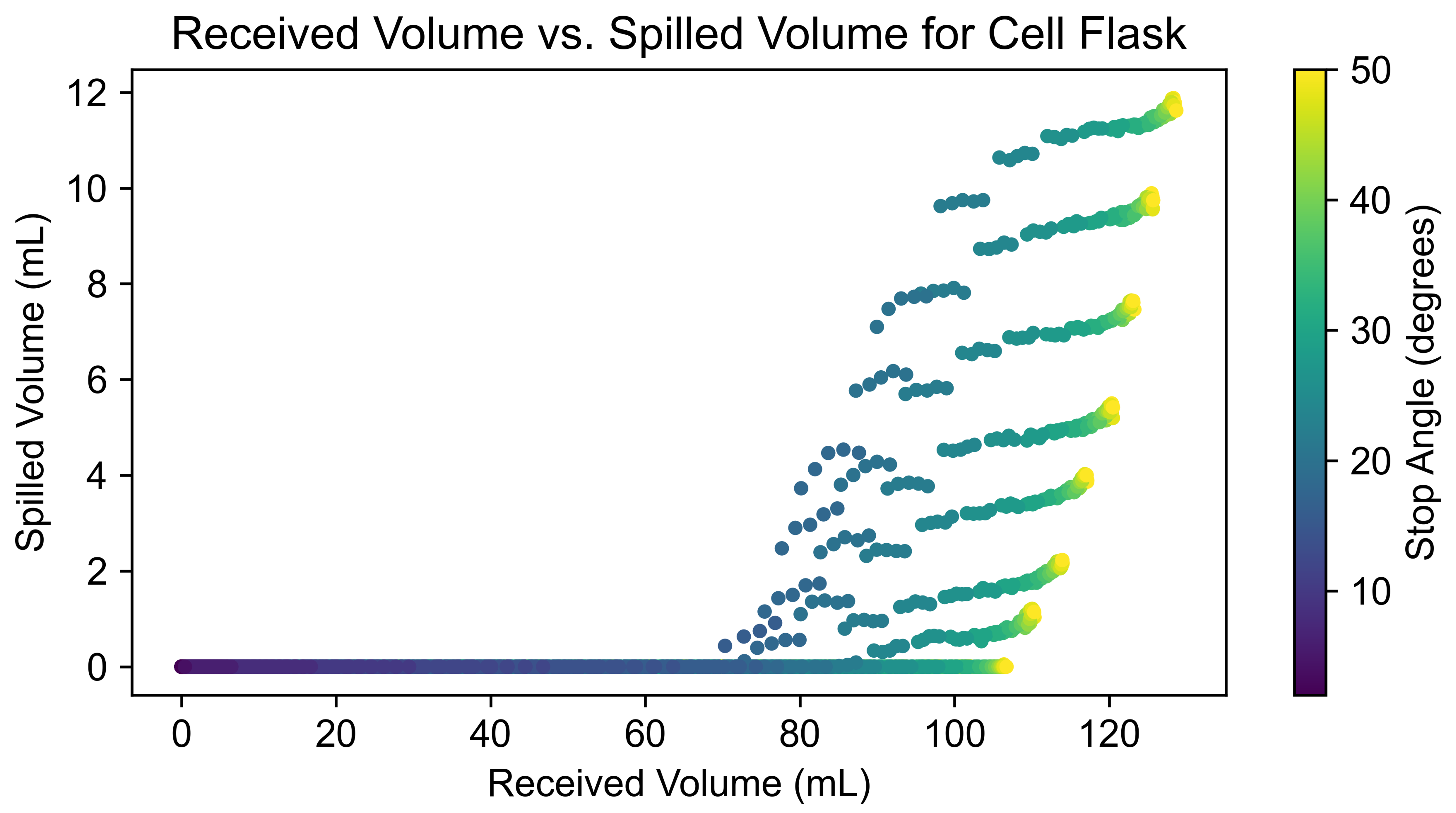}
  \caption{Visualization of the spilled volume over the received volume for the simulated pours with the cell culture flask. The color encoding shows the stop angle of the simulated pour.}
  \label{fig:Spilled_Volume_Flask}
\end{figure}

In Fig.~\ref{fig:Spilled_Volume_Flask}, we show the spilled volume in relation to the received volume and stop angle. The spilled volume increases for higher stop angles and received volumes. 

Additionally, it can be observed that the variation in the stop angle is responsible for the coarse differences in the poured volume, while the variation in the stop time results in more subtle differences. Similar results were found for the pouring simulation with the media bottle.

\subsubsection{Simulation-to-Reality}
\begin{figure}[!htb]
\centering
\includegraphics[width=1.0\linewidth]{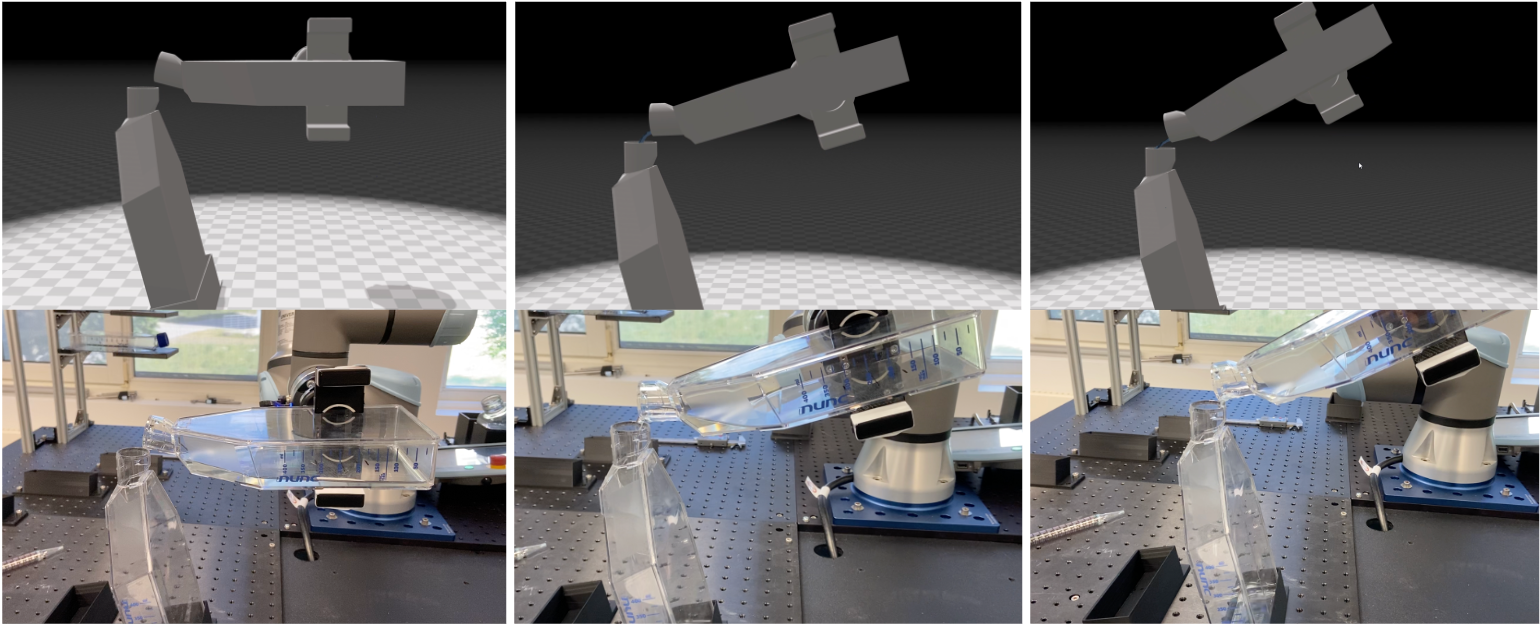}
\caption{A simulated pouring sequence with a cell culture flask (up) executed on the real-world robotic system (down).}
\label{fig:SimToRealitySeq}
\end{figure}

\begin{figure}[!htb]
\centering
\includegraphics[width=1.0\linewidth]{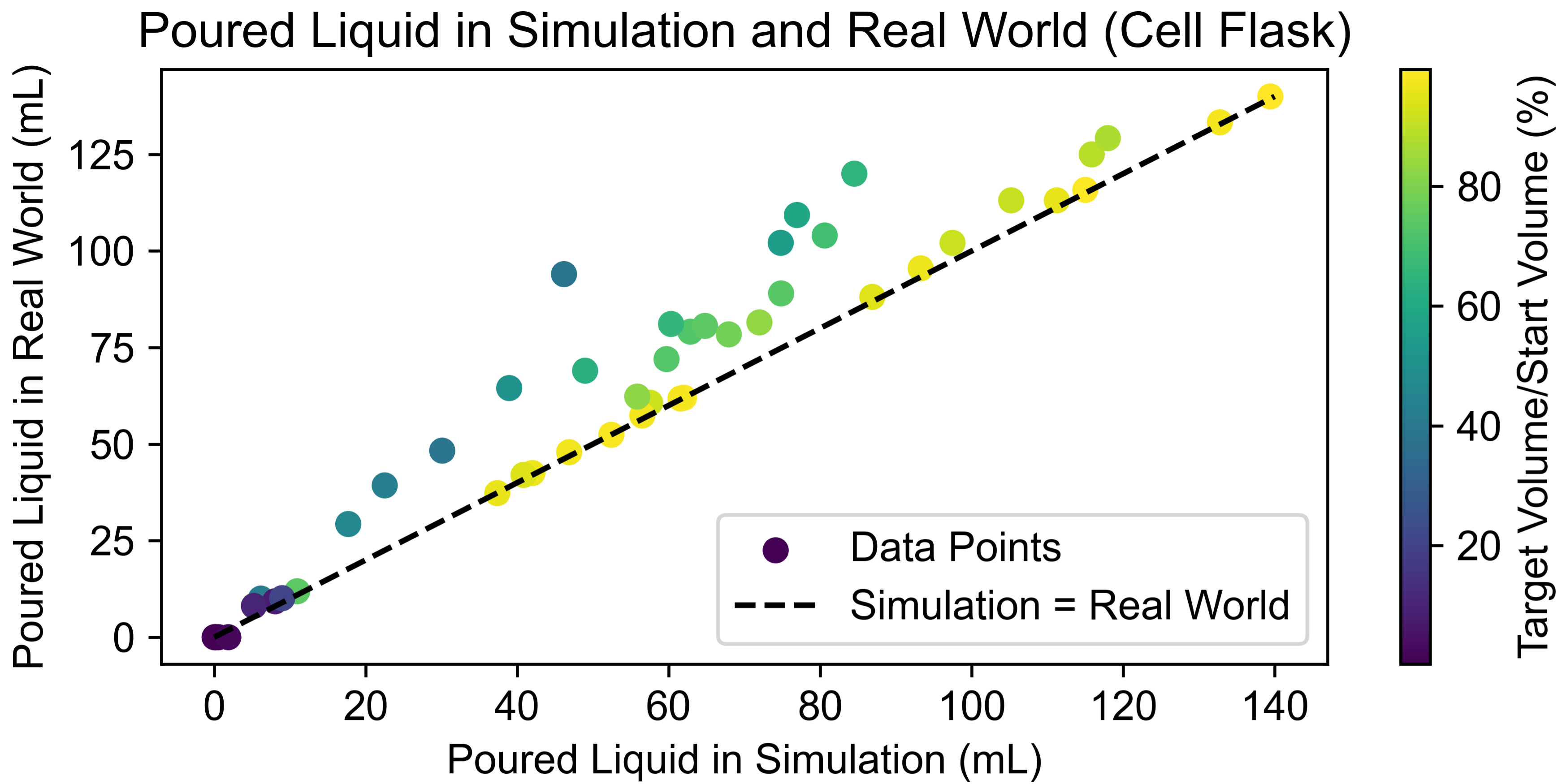}
\caption{Simulation-to-reality transfer results of the real experiments executed with a UR5e and a cell culture flask as pouring container.}
\label{fig:Sim_to_reality_Plot_Flask}
\end{figure}

The simulated movements were transferred to be executed on the system prototype using a UR5e as shown in Fig.~\ref{fig:SimToRealitySeq}. The results of 50 simulated pours executed in reality with the cell culture flask are visualized in Fig.~\ref{fig:Sim_to_reality_Plot_Flask}. The $RMSE$ is 10.8 mL ($MAPE$ = 21\%) with a maximum difference of 49.3 mL. The highest differences occur for ratios of the target volume and start volume between 0.3 and 0.7. For the media bottle, a $RMSE$ of 29.2 mL ($MAPE$ = 52\%) and a maximum difference of 52.6 mL was observed. For both containers, low spilling of on average less than 1 mL occurred during the real movements into a small opening.

\begin{figure}[!htbp]
  \centering
  \includegraphics[width=1\linewidth]{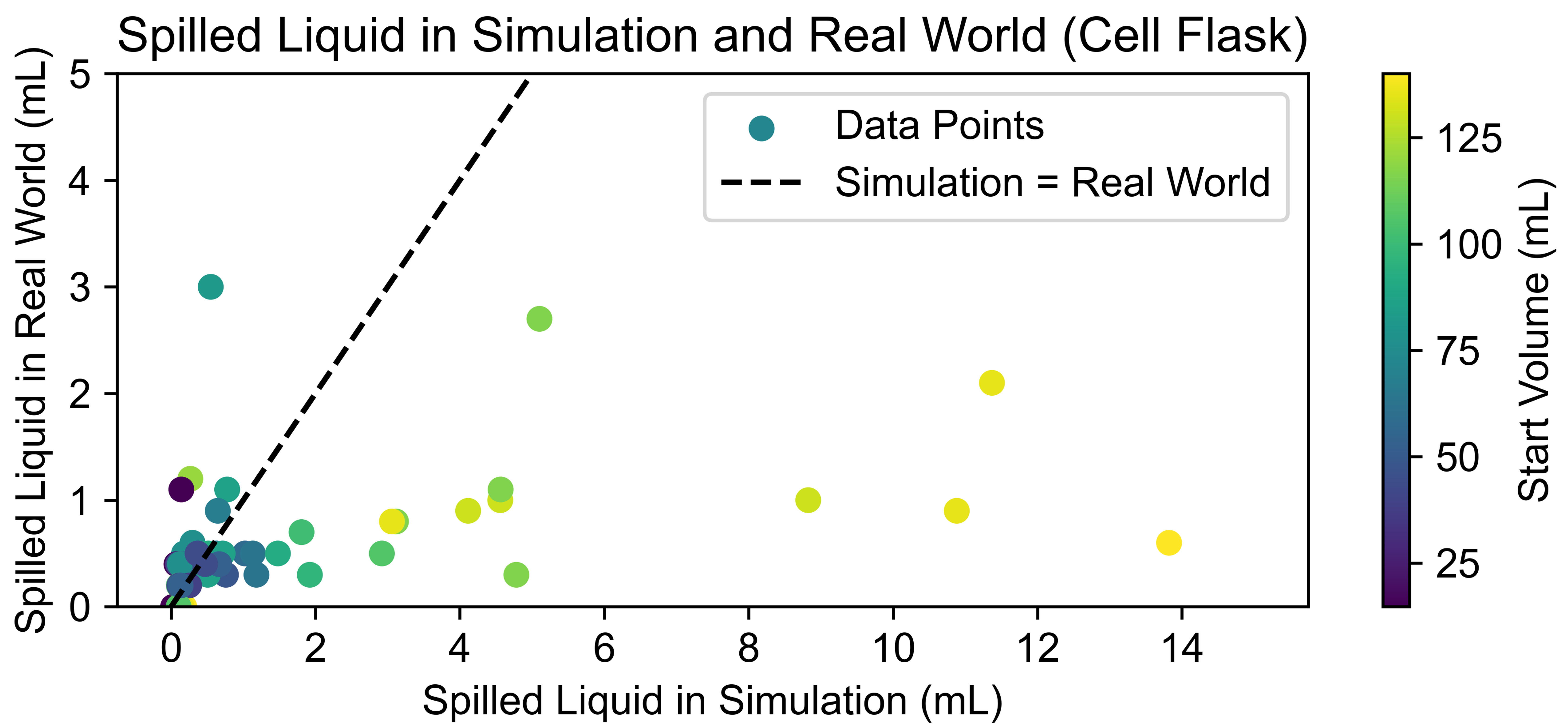}
  \caption{Spilled volume in the real world compared to the spilled volume in simulation. Again, we show the results for the cell culture flask as a pouring container.}
  \label{fig:Spilled_Volume_sim-to-real}
\end{figure}

The results of the spilled volumes in the real executions with the cell culture flask are visualized in Fig.~\ref{fig:Spilled_Volume_sim-to-real}. For 35 samples, the real spilled volume is lower than the predicted spilled volume from the simulation, with a maximum difference of 13.2 mL and a mean difference of 2.1 mL. For the remaining 15 pours, the real spilled volume is higher than the simulated one, with a maximum difference of 2.4 mL and a mean difference of 0.5 mL. It can be observed that the real spilled volume is close to the simulated spilled volume for low amounts of spilled liquid and low start volumes. 

\subsection{Overall Real-World Performance}
The precision of the autonomous pouring approach, which includes the vision-based volume estimation of liquids, the selection of the pour with the minimum cost from the simulation, and the real execution of the pour including picking up and returning the pouring container, was tested. Random start volumes ranging from 100 mL to 500 mL were used. The target volumes were specifically chosen to align with the typical amount of media poured by human scientists, which commonly falls within the range of 30 mL to 50 mL. The results are listed in Table \ref{tab:autonomous_pouring_results}. 

\begin{table}[!htb]
\small
\centering
\setlength{\tabcolsep}{8pt} % Adjusted column separation
\caption{Results of the complete autonomous pouring workflow tested on a UR5e (20 times per target volume).}
\begin{tabular}{@{}cccccc@{}}
\toprule
\makecell{Target\\Vol. (mL)} & \makecell{Avg. Exec.\\Time (s)} & \makecell{RMSE\\(mL)} & \makecell{MAPE\\(\%)} & \makecell{Max. Error\\(mL)} \\ \midrule
30                         & 135.3                        & 21.4                    & 71.3                    & 39.6                   \\
50                         & 124.8                        & 26.2                    & 52.4                    & 44.8                   \\ \bottomrule
\end{tabular}
\label{tab:autonomous_pouring_results}
\end{table}

The three main cell culture workflows (analyzing cell growth, changing media, passaging) were tested ten times each. The results are shown in Table \ref{tab:Workflows_Results}. Start conditions were one filled cell culture flask in the flask storage inside the incubator, five empty flasks in the flask storage on the table, a media and a washing solution bottle with sufficient liquid, and a filled trypsin container. Exemplary executions can be seen here: \url{https://github.com/DaniSchober/LabLiquidVision/tree/main/cell_culture_automation#--autonomous-workflows} .

\begin{table}[!htb]
\small
\centering
\caption{Results of the autonomous cell culture workflows executed on the proposed system (10 times each).}
\begin{tabular}{@{}ccc@{}}
\toprule
Workflow & \makecell{Avg. Exec.\\Time (s)} & \makecell{Completion\\Rate (\%)}\\ \midrule
Analyzing cell growth        & 78.3                                                                 & 100                  \\
Changing media        & 461.4                                                                & 100                  \\
Passaging        & 841.2                                                                & 90                   \\ \bottomrule
\end{tabular}
\label{tab:Workflows_Results}
\end{table}

\section{Conclusion}
In this work, we used a UR5 robot to automate the task of cell culturing. Our approach is independent of additional automation-specific hardware as well as the location of the pouring and receiving containers, which is especially useful for a flexible robotic laboratory assistant. In addition, using a pouring approach instead of pipetting reduces the amount of disposable plastics and avoids integrating non-automation-friendly pipettes. 
% When precise pouring is demanded, an option could include the vision-based monitoring of the liquid in the pouring and the receiving container.

We introduced a new pouring approach, which is especially relevant for receiving containers with small openings, leading to very low spilling of less than 1 mL on average.

For the vision-based liquid volume estimation, we show that incorporating depth maps alongside segmentation masks significantly enhances the liquid volume estimation model's performance, underscoring the benefits of utilizing depth information. We make the underlying dataset \textit{LabLiquidVolume} publicly available for the research community.

Our experimental results reveal that our setup is not suitable for precise laboratory experiments since the volume estimation and pouring simulation errors add up. However, improving the accuracy of both elements in the future could enable flexible automation of many research laboratory tasks.

\bibliographystyle{IEEEtran}
\bibliography{IEEEabrv, references_filtered}
\end{document}